%% file: paper.tex
\documentclass{article}
\usepackage{ijcai22}

\usepackage{times}
\usepackage{soul}
\usepackage{url}
\usepackage[hidelinks]{hyperref}
\usepackage[utf8]{inputenc}
\usepackage[small]{caption}
\usepackage{graphicx}
\usepackage{amsmath}
\usepackage{amsthm}
\usepackage{booktabs}
\usepackage{algorithm}
\usepackage{algorithmic}
\usepackage{makecell}
\usepackage{color}
\title{Psychiatric Scale Guided Risky Post Screening for Early 
Detection of Depression}

\author{
Zhiling Zhang \and
Siyuan Chen \and
Mengyue Wu$^*$ \And
Kenny Q. Zhu\thanks{Corresponding authors are supported by NSFC Grants No.91646205 and No.61901265, SJTU-CMBCC Joint Research Scheme, and Shanghai Municipal Science and Technology Major Project (2021SHZDZX0102).}
\affiliations
Shanghai Jiao Tong University
\emails
\{blmoistawinde, chensiyuan925, mengyuewu\}@sjtu.edu.cn,
kzhu@cs.sjtu.edu.cn
}

\begin{document}

\maketitle

\begin{abstract}
    Depression is a prominent health challenge to the world, and early risk detection (ERD) of depression from online posts can be a promising technique for combating the threat. Early depression detection faces the challenge of efficiently tackling streaming data, balancing the tradeoff between timeliness, accuracy and explainability. To tackle these challenges, we propose a psychiatric scale guided risky post screening method that can capture risky posts related to the dimensions defined in clinical depression scales, and providing interpretable diagnostic basis. A Hierarchical Attentional Network equipped with BERT (HAN-BERT) is proposed to further advance explainable predictions. For ERD, we propose an online algorithm based on an evolving queue of risky posts that can significantly reduce the number of model inferences to boost efficiency. Experiments show that our method outperforms the competitive feature-based and neural models under conventional depression detection settings, and achieves simultaneous improvement in both efficacy and efficiency for ERD.
\end{abstract}

\input{intro}
\input{method}
\input{experiment}
\input{related}

\input{conclusion}
\input{ethical}
\bibliographystyle{named}
\bibliography{paper}

\end{document}


\maketitle

\appendix

\section{Generalizability Tests}

Some studies have found that some existing depression detection models do not generalize well in the presence of distribution gaps among online-collected depression datasets \cite{harrigian2020models}. For instance, within 3 datasets collected from Reddit \cite{losada2016test,yates2017depression,wolohan2018detecting}, there are differences in the included subreddits, the time span of the posting histories, and the approaches to annotate the depressed users. As is suggested by \cite{ernala2019methodological}, current mental health prediction models tend to overfit on the characteristics of a specific dataset instead of learning what they claim to measure (i.e., a robust disease indicator). Therefore, even if there exists similarities for attempts in domain adaptation, the performance of current models still tend to degrade significantly. This highlights the difficulty for existing models to learn generalizable features. 

\subsection{Experimental Settings}

To test the generalizability of baseline models as well as the proposed method, we conduct experiments under conventional depression detection settings on 3 different datasets that have public availability and wide acceptance in previous works \cite{losada2017erisk,trotzek2018utilizing,harrigian2020models}. They are all collected from Reddit, but have different topic and label distribution and different content filtering strategy that aims to avoid label leakage and simulate user behaviors with different level of self-disclosure.\footnote{Although they label "depression" with different methods, they are all valid proxy signals of the same disease (``Depressive Disorder''), covering overlapping but slightly different subsets of patients. Therefore, the robustness across these depression subsets can indicate the general applicability of a method to some extent.} Besides eRisk2017, we introduce the two newly introduced datasets below. 

\paragraph{Topic-Restricted} consists of 6960 depressed users and 7683 control users, with a 8/1/1 random training/validation/test split \cite{wolohan2018detecting}. Since the original dataset is not released with the paper, we follow the implementation of \cite{harrigian2020models} to crawl the dataset ourselves in the same way. The depressed users are those who started a thread in depression subreddit while the control users are those who started a thread in AskReddit subreddit. The posting year spans from 2007 to 2021. For filtering, all posts in mental health related subreddits are removed. This stricter filtering strategy may further prevent model from overfitting on mental-health related signals, which can not be observed in depressed users who withhold their psychologic status due to the stigma of depression. 

\paragraph{RSDD} consists of 9210 depressed users and 107274 control users with a training/validation/test split of 39105/39122/39121, after data cleaning \cite{yates2017depression}. The depressed users are also identified with patterns, but further verified by annotators, while 12 control users are matched to a depressed user to minimize their distance of subreddit distribution. The posting year spans from 2006 to 2016. The filtering is the most strict among 3 datasets in that posts either posted in a mental health related subreddit or contain a depression-related term will be removed. This setting forces the model to learn the indirect signals for depression detection, so that they are more likely to detect the depression from patients with no self-report. However, it may also hinder the model's performance on those who would like to share their experience about depression.

For competing methods, we use the baselines described in \S 3, except HAN-GRU, BERT (Clus+Abs) for efficiency considerations. We additionally compare different variants of the proposed risky post screening strategy, including \textbf{Depress} using only 3 direct templates, \textbf{BDI-II} using 21 templates derived from BDI-II, and \textbf{Full} leveraging a combination of them (i.e. Psych described in \S 3). On the two new datasets, we find that a tiny version of BERT\footnote{https://huggingface.co/prajjwal1/bert-tiny} is enough to achieve competitive results given the larger data size. We select 64 posts, and train with batch size = 32 and learning rate = 2e-4. 

\subsection{In-domain Results}

\begin{table*}[t]
  \centering
\small
  \begin{tabular}{l|cccccc|c}
      \hline
          Source$\rightarrow$Target & T$\rightarrow$E & R$\rightarrow$E & E$\rightarrow$T & R$\rightarrow$T & E$\rightarrow$R & T$\rightarrow$R & Average  \\ 
          \hline
          LR & 82.6 & 72.3 & 68.8 & 67.3 & 77.8 & 52.0 & 70.1  \\
          Feature-rich & 85.7 & 75.3 & 69.6 & 73.3 & 77.6 & 52.7 & 72.4 \\ 
          \hline
          HAN-BERT (Depress) & 86.6 & 82.9 & 75.8 & 71.1 & 82.6 & \textbf{74.8} & 79.0  \\
          HAN-BERT (BDI-II) & \textbf{87.6} & 83.8 & 74.9 & \textbf{74.6} & 80.4 & 73.5 & 79.1  \\
          HAN-BERT (Full) & 87.4 & \textbf{85.0} & \textbf{77.5} & 72.4 & \textbf{84.4} & 72.3 & \textbf{79.8} \\ 
          \hline
      \end{tabular}
  \caption{Cross-domain experimental results (AUC) between eRisk2017(E), Topic-Restricted(T) and RSDD(R).}
  \label{table:cross} 
\end{table*}

\begin{table}[t]
  \centering
  \small
  \begin{tabular}{l|cc}
      \hline
      {} & Topic-Restricted & RSDD \\
      \hline
      LR & 69.8 & 52.1 \\
      Feature-Rich & 72.0 & 58 \\
      \hline
      BERT (Clus) & 56.7 & - \\
      HAN-BERT$_{tiny}$ (Heuristic) & 68.0 & 38.2 \\
      HAN-BERT$_{tiny}$ (Clus) & 71.9 & - \\
      \hline
      HAN-BERT$_{tiny}$ (Depress) & 77.1 & \textbf{65.4} \\
      HAN-BERT$_{tiny}$ (BDI-II) & \textbf{78.9} & 60.1 \\
      HAN-BERT$_{tiny}$ (Full) & 77.1 & 61.1 \\
      \hline
  \end{tabular}
  \caption{Test F1 on Topic-Restricted and RSDD dataset.}
  \label{table:rsdd_wolohan} 
\end{table}

The in-domain results on the 2 new datasets are shown in Table \ref{table:rsdd_wolohan}. BERT (Clus) and HAN-BERT (Clus) don't show competitive performance on Topic-Restricted again while requiring expensive clustering stage, so we don't experiment them on the larger RSDD dataset. The proposed risky post screening based methods show strong performance, and are capable of outperforming both the traditional Feature-Rich (all posts) using only 64 posts and a tiny version of BERT. The orders between these screening methods differ across datasets possibly due to their differences in label distribution and filtering strategy, but their performances are overall competitive.

\subsection{Cross-domain Results}

To test the cross-domain generalizability of different approaches, we train models on a source dataset and directly test the model on another target dataset for all 6 possible combinations of the 3 datasets. For HAN-BERT models, we use the model trained with 16 selected posts, and also test on 16 posts selected with the same templates. In contrast to in-domain experiments using a fixed probability threshold 0.5 to decide the prediction, we don't apply this in cross-domain tests since the label distribution differs greatly between datasets, so that fixed threshold will lead to poor performance. Instead, we use a threshold-free metric, AUC, to measure the performance of each method.

The results are shown in Table \ref{table:cross}. Consistent with in-domain results, we find that Feature-Rich outperforms LR in terms of average domain adaptation AUC. Moreover, the performance of all HAN-BERT models are more robust than baselines, which suggests the generalizability of the proposed method.

We then analyze the results in detail. The performance of LR and Feature-rich degrade significantly in the T$\rightarrow$R setting. We hypothesize that the reason lies in the different annotation strategy for depressed and control users. Topic-Restricted treats users who started a thread in depression/AskReddit subreddit as depressed/control users and does not control the similarities between the 2 groups, while RSDD specifically selects control users with similar subreddit distribution as depressed users. Therefore, the baselines may have learned spurious clues about the annotation strategy of Topic-Restricted. For example, we checked the coefficients of the LR model, and found that words like ``askreddit'' and ``redditors'' are among the most important features for the decision of control users. In contrast, HAN-BERT models still exhibit satisfying performance, which suggests that they can leverage robust depression indicators. 

LR and Feature-Rich perform much worse than HAN-BERT models in both direction between RSDD and eRisk2017. These two datasets adopts different filtering strategies, where eRisk2017 preserves most posts while RSDD excludes all posts in mental health related subreddits or containing a depression-related term. The results indicate that HAN-BERT models are less affected by such domain gap. 

HAN-BERT models also significantly outperform baselines in the E$\rightarrow$T and E$\rightarrow$R settings, which shows that they can effectively capture depression signals even with the extremely small eRisk2017 dataset. Comparing all variants of HAN-BERT models, HAN-BERT (Full) shows the best average performance, which indicates the usefulness of combining the theory-guided templates with direct depression indicators in selecting robust features across domains. Therefore, HAN-BERT (Full) can be a preferred choice in real world applications where the target domain distribution is unknown, and we select this variant as the representative of the proposed method.

\section{Lexical Analysis}

\begin{figure*}[t]
    \centering
    \includegraphics[width=2\columnwidth]{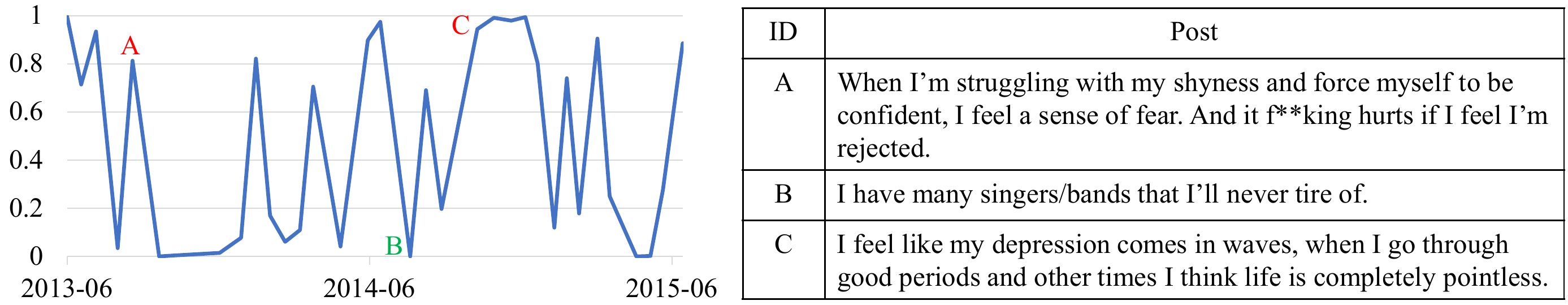}
    \caption{Predicted depression score by HAN-BERT (Full) along with time for a user in eRisk2017 dataset. We selected 3 time periods on predicted peaks and troughs, and show a representative post in each period.}
    \label{fig:curve}
\end{figure*}

As has been shown in many previous works \cite{shen2017depression,eichstaedt2018facebook,wolohan2018detecting}, there are many significant lexical differences between the posts of depressed users and other users, which can be captured by the word frequency of certain categories in LIWC \cite{pennebaker2001linguistic}. For example, depressed users tend to use more \textbf{first person pronouns} (\textit{I}), words expressing \textbf{negative emotions} (e.g. \textit{hate, miss, alone}), and words about \textbf{health} (e.g. \textit{life, tired, sick}). Such lexical discrepancies do not only exist between the two groups of users, but also within the posts of depressed users themselves. Therefore, we run risky post screening on the posts of depressed users in the eRisk2017 test set, and count the frequency of words in the 3 LIWC categories stated above. We then compare the proportion of these words in selected posts and other posts, and test their differences with two-sided proportion $z$ test. If the selected posts show stronger lexical depression indicators, we can further confirm the helpfulness of risky post screening in capturing reliable depression features.

\begin{figure}[h]
  \centering
  \includegraphics[width=0.88\columnwidth]{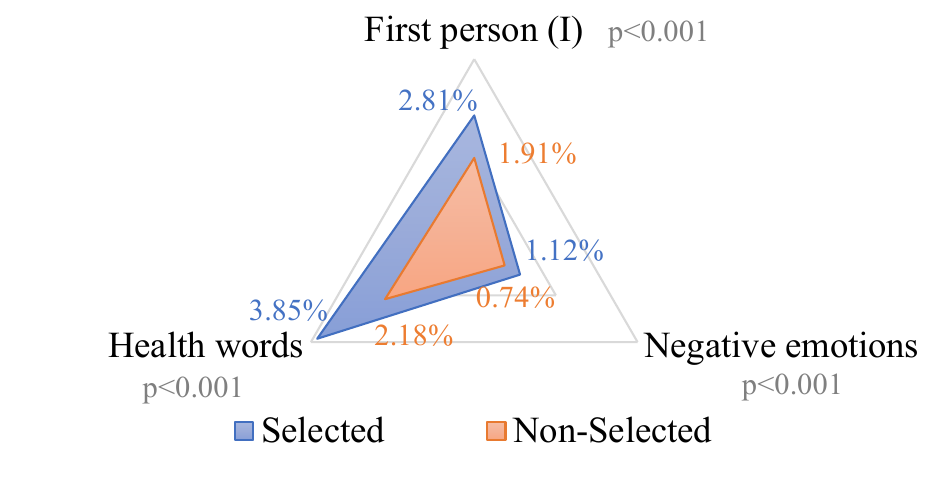}
  \caption{Difference between selected and non-selected posts w.r.t. proportion of words in typical depress related LIWC categories.}
  \label{fig:lexical}
\end{figure}

As is illustrated in Figure \ref{fig:lexical}, there are significant differences between the use of first person words, negative emotions and health-related words in selected (risky) and non-selected posts ($p < 0.001$ for all categories). The difference between risky posts and other posts of depressed users (Negative Emotion: 1.12\% vs. 0.74\%) can be even bigger than the difference between non-risky posts of depressed users and all posts from non-depressed users (Negative Emotion: 0.74\% vs. 0.79\%). These findings are similar to those reported in previous literature, which verifies the convergent validity of the proposed method.

\section{Temporal Analysis}

We also demonstrate that our method has the potential to track the fluctuations in depressive mood for depression patients by another example (Figure \ref{fig:curve}). We produce such curve with the following procedure. First, we group posts according to a 14-day interval. Then we use HAN-BERT (Full) to conduct post screening and depression detection to get a predicted probability for each post group. To produce a smooth curve along time, we design a moving average strategy to derive a more stable depression score from predicted depression probabilities. Suppose the predicted probability of group $i$ is $pr_i$, and the depression score is $s_i$. Then we have 
$$s_1 = pr_1, s_i = \alpha s_{i-1} + (1-\alpha) pr_{i-1} (i > 1)$$
where $\alpha = max(0, 0.5 * \frac{28-t_{\Delta}}{28-1})$, $t_{\Delta}$ is the time interval between the first post of two groups measured in days. Therefore, if two groups are close in time, then the score from the last period will have a higher influence on the next score. Finally, we plot the curve according to the moving-averaged scores.

As we can see in Figure \ref{fig:curve}. In addition to the accurate detection of depression, the proposed method may also be able to capture the changes in the severity of depression symptoms. The model reported a high risk when the user expressed typical depression symptoms like frustrations (A) or worthlessness (C), and also reported low depression score when the user actually showed interest in things (B), which might indicate a recovery from the symptom of ``Loss of Interest''. The overall trend is in line with the user's self-report that the depression comes cyclically like waves.

\section{Hyperparameter Analysis}
\label{sec:factor}

Here we study the impact of 2 hyperparameters of risky post screening.

\begin{figure}[h]
    \centering
    \includegraphics[width=0.8\columnwidth]{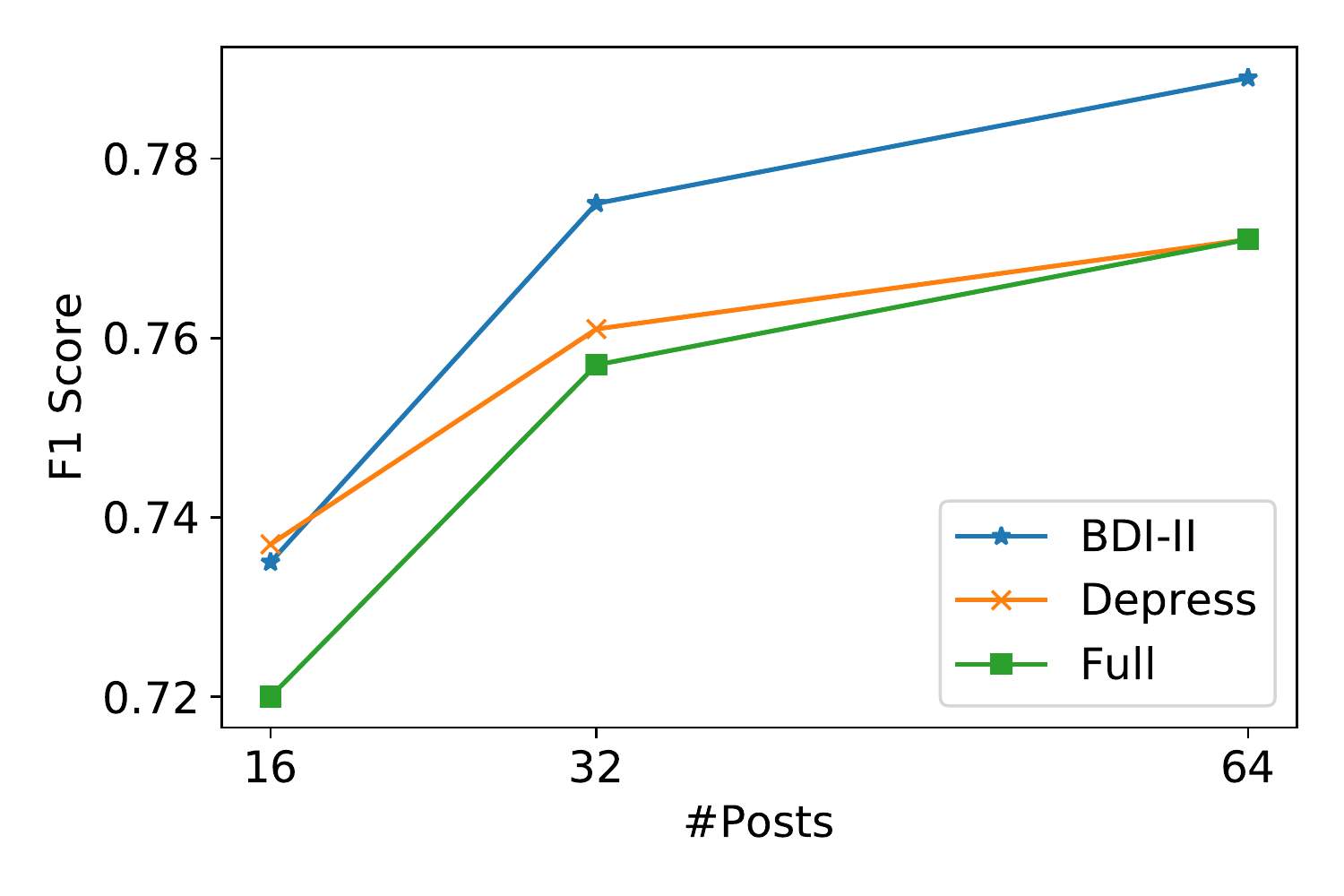}
    \caption{Effect of number of posts on Topic-Restricted.}
    \label{fig:post_numbers}
\end{figure}

\paragraph{Number of Posts} 
We study the effect of post numbers on the Topic-Restricted dataset (Figure \ref{fig:post_numbers}). It can be seen that all methods can get further improvement given more posts, and scale-based methods can benefit more possibly because more posts can help cover more diverse expressions of depression symptoms, which cannot be fully captured given a small size limit.

\paragraph{Different Scales}
We explore three other commonly-adopted scales besides BDI-II, including HDRS \cite{hamilton1986hamilton}, CES-D \cite{Lenore1977CES-D} and PHQ-9 \cite{kroenke2001phq}, with similar approaches to rewrite the dimensions into depression templates. We also tried Majority Voting of models paired with different scales. Their performance on eRisk2017 dataset in shown in Table \ref{tab:scales}. We found that BDI-II is the single best performing scale. The combination of HDRS, BDI-II and PHQ-9 reaches the highest performance for their highly complementary dimensions. 

\begin{table}[h]
    \centering
	\small
    \begin{tabular}{l|c}
        \hline
        Depression Scale & F1 \\
        \hline
        BDI-II & \textbf{70.3} \\
        HDRS & 68.0 \\
        CES-D & 67.9 \\
        PHQ-9 & 67.2\\
        \hline
        HDRS+BDI-II+PHQ-9 & \textbf{72.1} \\
        HDRS+BDI-II+CES-D & 70.5 \\
        \hline
    \end{tabular}
    \caption{Test results on eRisk2017 dataset with different depression scales and their combinations.}
    \label{tab:scales}
\end{table}

\section{Depression Templates}

Here we provide the templates in detail. We mainly use a combination of 3 direct depression descriptions and the 21 indirect symptoms derived from BDI-II (Table \ref{tab:BDI-II}) \cite{beck1996beck}. As is mentioned in \S \ref{sec:factor}, we also experimented other well-known depression scales like HDRS (Table \ref{tab:Hamilton}) \cite{hamilton1986hamilton}, CES-D (Table \ref{tab:CES-D}) \cite{Lenore1977CES-D} and PHQ-9 (Table \ref{tab:PHQ-9}) \cite{kroenke2001phq}. The original scales usually contain different descriptions under the same dimension to distinguish different level of intensity or frequency. However, we find that current sentence representations have difficulty in capturing such nuanced differences. We thus condense the descriptions of each dimension into one general template (A few may have more, if there are significant intra-dimension difference).

\begin{table}
  \centering
  \small
  \begin{tabular}{l|l}
  \hline
  Dimension & Template \\
  \hline
  Feeling Depressed  &  I feel depressed. \\
  Diagnosis &  I am diagnosed with depression. \\
  Treatment &  I am treating my depression. \\
  \hline
  Sadness & I feel sad.  \\
  Pessimism & I am discouraged about my future.  \\
  Past Failure & I always fail. \\
  Loss of Pleasure & I don't get pleasure from things. \\
  Guilty Feelings & I feel quite guilty. \\
  Punishment Feelings & I expected to be punished. \\
  Self-Dislike & I am disappointed in myself. \\
  Self-Criticalness & I always criticize myself for my faults. \\
  Suicidal Thoughts or Wishes & I have thoughts of killing myself. \\
  Crying & I always cry. \\
  Agitation & I am hard to stay still. \\
  Loss of Interest & It's hard to get interested in things. \\
  Indecisiveness & I have trouble making decisions. \\
  Worthlessness & I feel worthless. \\
  Loss of Energy & I don't have energy to do things. \\
  Changes in Sleeping Pattern & I have changes in my sleeping pattern. \\
  Irritability & I am always irritable. \\
  Changes in Appetite & I have changes in my appetite. \\
  Concentration Difficulty & I feel hard to concentrate on things. \\
  Tiredness  & I am too tired to do things. \\
  Loss of Interest in Sex & I have lost my interest in sex. \\
  \hline
  \end{tabular}
  \caption{The main templates and their corresponding dimensions we used in our experiments, including 3 direct depression descriptions and 21 indirect symptoms derived from BDI-II. }
  \label{tab:BDI-II} 
\end{table}

\begin{table}
    \centering
    \small
    \begin{tabular}{l}
      \hline
      I have depressed mood. \\
      I always feel sad. \\
      I feel hopeless. \\
      I feel helpless. \\
      I find myself worthless. \\
      I have feelings of guilty. \\
      I always let people down. \\
      I feel like I should be punished. \\
      I think life is not worth living. \\
      I have thoughts of killing myself. \\
      I tried to suicide. \\
      I have difficulty falling asleep. \\
      I feel restless. \\
      I always wake up during the night. \\
      I have lost my interest in many things. \\
      I decrease time spent in my job. \\
      I find it difficult to concentrate on things. \\
      I can not stay still. \\
      I always worry about small things. \\
      I am irritable. \\
      I feel anxiety. \\
      I have a bad appetite. \\
      I am easy to be tired. \\
      I have less interest in sex. \\
      I suffers from menstrual disturbances. \\
      I worry about my health. \\
      I lose weight dramatically. \\
      \hline
    \end{tabular}
    \caption{The templates adapted from the HDRS depression scale. }
    \label{tab:Hamilton} 
  \end{table}
  
  \begin{table}
    \centering
    \small
    \begin{tabular}{l}
      \hline
      I am bothered by things that usually don't bother me. \\
      I do not feel like eating. \\
      My appetite is poor. \\
      I feel that I could not shake off the blues even with help \\
      -from my family or friends. \\
      I am not just as good as other people. \\
      I have trouble keeping my mind on what I am doing. \\
      I feel depressed. \\
      I feel that everything I did was an effort. \\
      I feel hopeless about the future. \\
      I thought my life had been a failure. \\
      I feel fearful. \\
      My sleep is restless. \\
      I am unhappy. \\
      I talk less than usual. \\
      I feel lonely. \\
      I think people are unfriendly. \\
      It's difficult for me to enjoy life. \\
      I had crying spells. \\
      I feel sad. \\
      I feel that people dislike me. \\
      I could not get 'going'. \\
      \hline
    \end{tabular}
    \caption{The templates adapted from the CES-D depression scale. }
    \label{tab:CES-D} 
  \end{table}
  
  \begin{table}
    \centering
    \small
    \begin{tabular}{l}
      \hline
      I have little interest in doing things. \\
      I have little pleasure in doing things. \\
      I always feel down. \\
      I always depressed. \\
      I always hopeless. \\
      I have trouble falling asleep. \\
      I sleep too much. \\
      I feel tired. \\
      I have little energy. \\
      My appetite is poor. \\
      I cannot stop overeating. \\
      I feel bad about myself. \\
      I think myself a failure. \\
      I have let other people down. \\
      I have trouble concentrating on things. \\
      I move much slower than before. \\
      I speak much slower than before. \\
      I have been moving around a lot more than usual. \\
      I think that I would be better off dead. \\
      I have thoughts of hurting myself. \\
      \hline
    \end{tabular}
    \caption{The templates adapted from the PHQ-9 depression scale. }
    \label{tab:PHQ-9} 
  \end{table}

\section{Ethical and Broader Impact Statement}

This work aims to help people suffering from depression, but have not yet been diagnosed due to the difficulty in receiving clinical help or the stigmatization of the disease. It can be a sensitive topic so it is important to discuss the potential risks and limitations of our work. The proposed method can conduct early depression detection on social media. However, the performance is far from prefect, so the models' early alerts still require careful examinations from professional practitioners. The proposed method can provide diagnostic bases as explanations, but the diagnostic basis may not precisely matched the actual symptom implied in the post. Therefore, the diagnostic basis should be checked before adoption. Moreover, the datasets are annotated with proxy signals of depression, which may not be representative of the true population of depression patients. In practice, the model should be trained on a more carefully-curated dataset for reliable predictions.

The datasets used in this work are either publicly available or used under their corresponding data usage agreement. All posts in examples were de-identified and paraphrased for anonymity.

\bibliographystyle{named}
\bibliography{paper}

%% file: intro.tex
\section{Introduction}

Depression has been a major health challenge to the world, with over 280 million people affected, according to WHO\footnote{\url{https://www.who.int/news-room/fact-sheets/detail/depression}}. Moreover, the COVID-19 pandemic has further deteriorated the situation. 
Since people are more 
willing to express their feelings on online social media during this 
special period\footnote{\url{https://www.statista.com/statistics/1106498/home-media-consumption-coronavirus-worldwide-by-country/}}, 
depression detection from online posts can be a promising approach 
to combat the challenge.
The conventional setting of depression detection from online posts is to predict whether a user suffers from depression from the whole posting history \cite{gui2019cooperative,zogan2021depressionnet}. However, for social networks which update quickly, 
another setting, \textit{early risk detection} (ERD)~\cite{losada2017erisk}, 
may have more potential to detect and offer timely help to risky users. 
An ERD model should access user post one by one sequentially, 
dynamically update the estimated risk, and make immediate alert once 
it is confident enough about its prediction. This setting is less explored due to 
its unique challenges: 

First, the ability of classification on streaming data is a requirement of ERD models. This means that the method is better to be an online, incremental algorithm that can update the prediction every time a user sends a post, rather than an offline batch algorithm that only runs once after a long interval. Since traditional ML models do not 
come with such ability inherently, a typical solution is to naively process 
the whole posting history for each update~\cite{trotzek2018utilizing}. 
This method can hardly be efficient enough in practice. 
For instance, many systems in an ERD competition, eRisk2019, spent several 
days for computation~\cite{losada2019overview}. 

Moreover, an ERD model should make tradeoff between its timeliness and accuracy. 
To make an early prediction, the model usually predicts a depression probability 
after each update, and makes an alert if the probability exceeds certain threshold, and we can tune the threshold to control the latency of prediction \cite{trotzek2018utilizing}. Leveraging more posts can certainly facilitate higher accuracy, while it also means that the model makes late predictions, and it fails to make alert before the patient's condition deteriorates. To realize the pareto improvement of both objectives, 
we should also seek improved model structure. 
Although large pretrained Language models (LMs) like BERT~\cite{devlin2018bert} 
has achieved great success in many classification tasks, 
they are seldom applied to ERD, as the long posting history make 
the memory cost and latency prohibitive. 

Due to the sensitiveness of depression detection, model explainability is also 
a vital property. Without proper explanations, it can be hard for users to trust such novel tools and accept these alerts. Since 
deep learning models are mostly black-boxes, one cannot ascertain whether their
predictions are achieved due to robust features, or some spurious clues~\cite{ribeiro2020beyond}. Traditional ML models can provide global explanations of the prediction (i.e., feature importance) based on features like word counts. However, it is much
more preferable if we can make personalized, symptom-based explanations~\cite{mowery2017understanding} like psychiatrists to endow higher level of trustworthiness. 

\begin{figure}[t]
    \centering
    \includegraphics[width=\columnwidth]{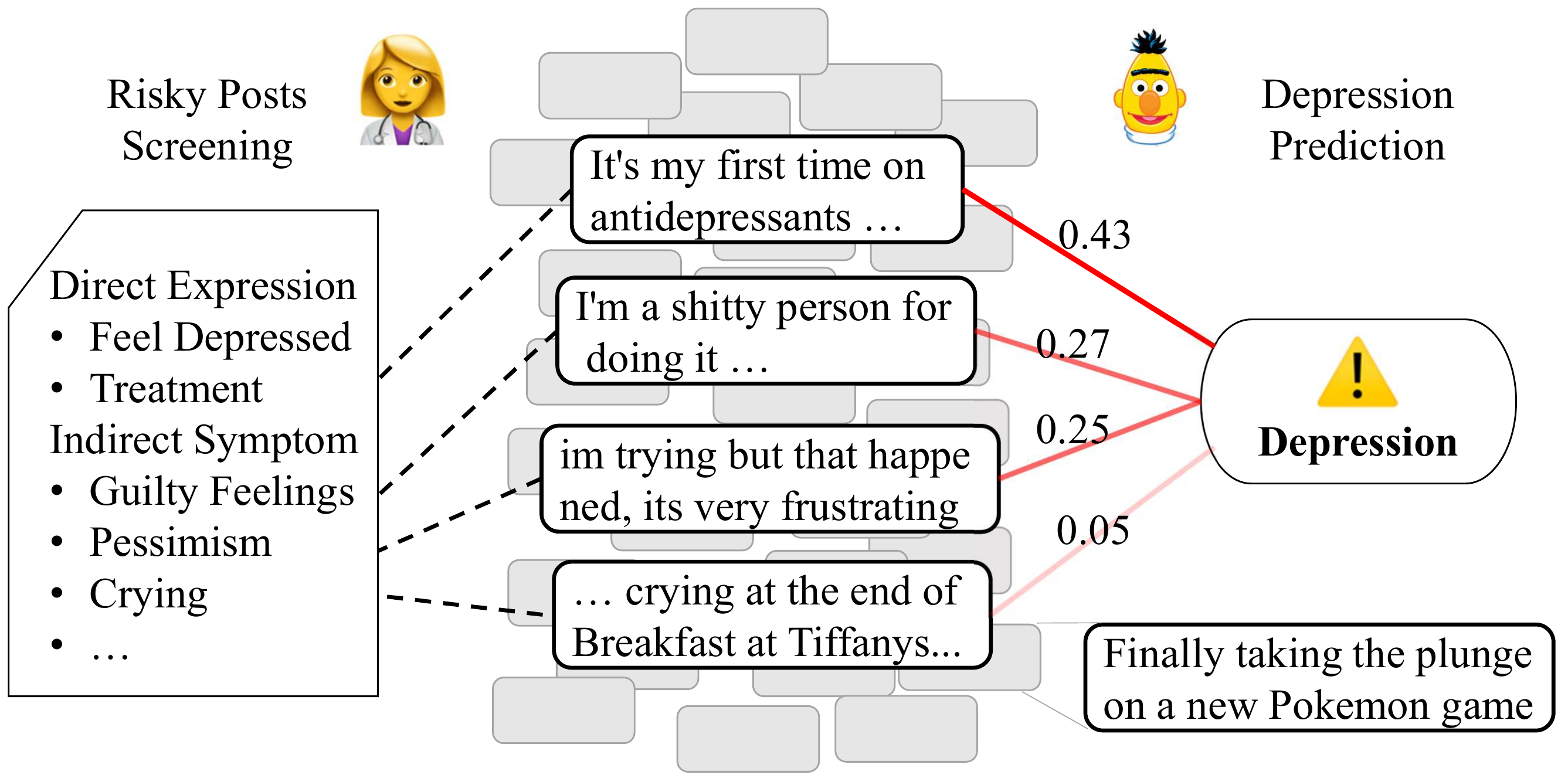}
    \caption{Overview of the system. Depression templates derived from established depression scales are used to screen risky posts, and filter out safer ones (boxes in grey). A hierarchical attentional network further attends to truly important contents (darkness of red lines indicates attention strength) and makes final prediction. }
    \label{fig:overview}
\end{figure}

Inspired by the psychiatry practice of using clinical scales to screen depression patients, we propose to use depression templates derived from established depression scale \cite{beck1996beck} to screen risky posts. These templates include direct expressions of depressive moods and depression treatments, as well as theory-grounding indirect symptoms like guilty feelings, pessimism and loss of appetite, etc. Only posts highly relevant to these templates will be selected out of the whole posting history, which can greatly reduce the input size, so as to eliminate distractors and improve processing efficiency. A hierarchical network incorporating attention mechanism \cite{yang2016hierarchical} and BERT \cite{devlin2018bert} further aggregates the selected posts of a user, and assigns higher weights to truly important contents for accurate and explainable predictions. The overview of our approach is illustrated in Figure \ref{fig:overview}. To enable ERD, we also propose an online algorithm based on a risky posts queue evolving with the streaming posts. Experimental results show that the proposed method can achieve SOTA performance in both conventional and ERD settings, and can be even a more efficient ERD solution than simple Logistic Regression models. \footnote{Code, Appendix and the used scale data at: \url{https://github.com/blmoistawinde/scale_early_depress_detect}} 

Our key contributions are as follows: 
1) We propose psychiatry-guided risky post screening to select salient contents for processing, which reduces the input size so as to allow the utilization of large models, and can provide symptom-based interpretations.
2) We leverage hierarchical attentional network with BERT (HAN-BERT) to enhance the model accuracy and explainability.
3) We propose an online algorithm based on an evolving queue of risky posts to tackle ERD, achieving simultaneous improvement in timeliness and accuracy over representative baselines. 

%% file: method.tex
\section{Methods}

For a user $U_i$ with posts $[P_{i,1}, P_{i,2}, ..., P_{i,n}]$ in the activity history, where $n$ is the number of total posts and $P_{i,j}$ is the $j$-th user-generated post of $U_i$, the goal of \textit{conventional depression detection} is to predict a binary label $y_i \in \{0, 1\}$ indicating whether the user $U_i$ suffers from depression, given the whole activity history. In contrast, in an \textit{early risk detection} (ERD) setting, the posts come one by one, so that only $[P_{i,1}, P_{i,2}, ..., P_{i,t}]$ is available to the model at the $t$-th time. The model can make an early prediction of $y_i$ at $t (t \leq n)$ once it is confident enough, such that the prediction can make a good tradeoff between accuracy and earliness (with $t$ as small as possible). Our solutions for both settings are as follows. 

\subsection{Risky Post Screening}
\label{sec:screening}

A reddit user typically has hundreds or thousands of posts in the whole activity history. However, since not all user posts are relevant to the detection of depression, retaining all posts may run the risk of introducing distractors that can hinder model performance and efficiency. Therefore, effective post selection strategy can be crucial to the success of ERD. 

Intuitively, posts that directly disclose the state of depression or express depression-related symptoms would indicate high depression risks. These intuitions can be reliably captured by psychometrically validated clinical scales. Therefore, we draw inspiration from these scales, and devise \textbf{depression templates} to screen risky posts out of the lengthy activity history. Only the posts with highest similarities to these templates will be selected as risky posts for further prediction.

Our depression templates are made up of 2 groups of descriptions. 
The first group consists of 3 explicit depression-related expressions: 
``I feel depressed'', ``I am diagnosed with depression'', 
``I am treating my depression'', matching the person's claim of 
general depressive mood, the diagnosis and the post-diagnosis treatment,
respectively. The second group is comprised of descriptions corresponding 
to dimensions defined in a clinical depression scale. 
Here we mainly adopt BDI-II, which is one of the most widely used depression 
measures\footnote{We also experimented with other scales and combinations of multiple scales, see Appendix for more details.} \cite{beck1996beck}. 
The scale includes the descriptions of four different intensities for each of the 21 symptoms. For example, ``1: I do not feel sad'', ``2: I feel sad much of the time'', ``3: I am sad all the time'' and ``4: I am so sad or unhappy that I can't stand it'' for the symptom of sadness. However, we find that current sentence representations have difficulty in capturing such nuanced differences. Therefore, we make slight manual modifications on them to construct a single representative template for each dimension, like ``I feel sad'' for Sadness. We provide examples in Table \ref{tab:template_example}.

\begin{table}[h]
    \centering
    \begin{tabular}{l|l}
        \hline
        Dimension & Template \\
        \hline
        Crying & I always cry. \\
        Tiredness  & I am too tired to do things. \\
        Self-Dislike & I am disappointed in myself. \\
        \hline
    \end{tabular}
    \caption{Example BDI-II dimensions and their corresponding templates.}
    \label{tab:template_example}
\end{table}

To measure the similarity between posts and depression templates, we resort to pretrained sentence-transformers \cite{reimers-2019-sentence-bert} to get the sentence representations, and calculate the cosine similarity between each post-template pair. For a post, its most similar template is referred to as its \textbf{diagnostic basis}, and their similarity is regarded as the \textbf{risk} of the post. The process of \textbf{risky post selection} is to select at most $K$ posts with the highest risks out of all posts of a user. Since the procedure consists of only sentence encoding from a pretrained model, and cosine similarity calculations, our method can be more efficient than previous works on post selection that requires costly RL training \cite{gui2019cooperative}. Moreover, the theoretical underpinning basis for post selection is also more effective than heuristic-based selection such as clustering \cite{zogan2021depressionnet}, as we will validate in the experiments (\S \ref{sec:conventional}).

\subsection{Hierarchical Attentional Network}
\label{sec:HAN}

Although depression detection can be formulated as text classification problem, it is different from conventional settings in that the input consists of multiple posts attached with temporal information. We may reformat the input by simply concatenating all posts. However, such representations will lost the time and structural clues at the post level, and also lead to a lengthy sequence. Further, conventional text classification models are lacking in explainability. 

To leverage the posting list structure as well as providing post-level explainability, we adopt the framework of Hierarchical Attentional Network (HAN) \cite{yang2016hierarchical} in our model design. The HAN consists of a post encoder and a user encoder. 

The post encoder takes the words $\{x_1, x_2, ..., x_L\}$ in a single post, and encode them into a post representation $p$. Thanks to the pre-step of risky post screening, we are able to take advantage of a large post encoder as opposed to shallow CNN or GRU based structures used in previous works \cite{yates2017depression,zogan2021explainable}. Hence we use a pretrained BERT model as the post encoder and the representation of the [CLS] token as the representation for the whole post. Therefore, the post encoder can be represented as: 

\begin{equation}
    p = BERT_{[CLS]}(x_1, x_2, ..., x_L)
\end{equation}

Given the representations of the $K$ risky posts $\{p_1, p_2, ..., p_K\}$, the user encoder models the relations between these posts as well as their chronological order to produce updated contextualized representations of each post $\{p'_1, p'_2, ..., p'_K\}$, and further aggregate these embeddings into one user representation $u$. Here we utilize a transformer structure to model the posts' relations with self-attention and encode the order with positional embeddings. The updated post representations are further passed to an attentional pooling layer, which learns the weight for each post embedding and perform a weighted sum of them accordingly, to get the final user representation. The attention mechanism can distinguish the contributions of each posts with learned weight so that important posts will have a higher influence on the final prediction. After all, the user encoder can be represented as:

\begin{equation}
    p'_1, p'_2, ..., p'_K = Transformer(p_1, p_2, ..., p_K)
\end{equation}
\begin{equation}
    \alpha_k = \frac{exp(W p'_k + b)}{\sum_{k'=1}^{K} exp(W p'_{k'} + b)}
\end{equation}
\begin{equation}
    u = \sum_{k=1}^K \alpha_k p'_k
\end{equation}
where $W$ and $b$ is learnable weight matrix and bias term of the linear transformation in the attentional pooling layer. Finally, a linear layer on top of the user representation makes the final binary classification of depression. The whole model is trained with the standard binary cross entropy loss.

With the HAN model above, most of the single post will not exceed BERT length limit. The attention weights can also provide explanations for which post is considered vital in the model's decision. 

\begin{figure*}[htbp]
    \centering
    \includegraphics[width=1.5\columnwidth]{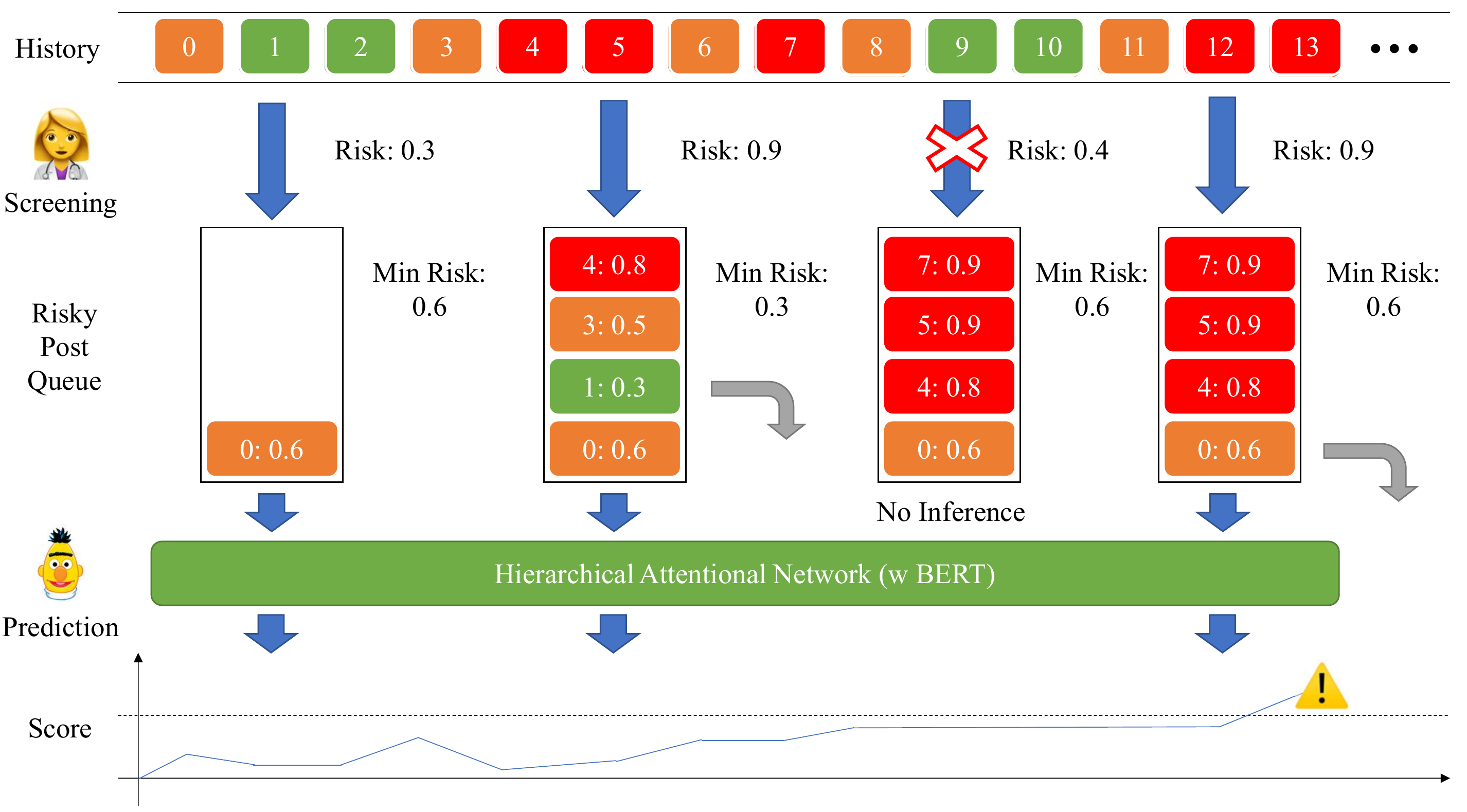}
    \caption{Illustration of the evolving queue for early risk detection.}
    \label{fig:evolving}
\end{figure*}

\subsection{Evolving Queue for Early Detection}
\label{sec:evolving}

In an ERD scenario, we need to incrementally make predictions each time a user posts, instead of processing the whole posting history once.
This brings the computational challenges of frequent \textit{feature updates} and \textit{model inferences}. For example, a traditional feature-based model would have to recalculate the features like LDA topic distribution \cite{blei2003latent} from the whole activity history after each single update, and then make prediction accordingly. Such frequent recalculations can be even intractable for typical DNN solutions. Although post selection strategies can reduce the computational costs at the \textit{model inference} stage to some extent, the selection stage itself can become a performance bottleneck if it is not efficient enough. 

Moreover, the frequent updates may also be sensitive to one single depression-like post and easily produce false positive predictions, while the depressed patients tend to suffer from durative symptoms  \cite{kroenke2001phq} as opposed to control users who can also be depressive at some moments. Since the ERD setting does not allow modifying the prediction, once the model makes a confident diagnosis (since we may have already taken action to intervene in the dangerous situation), these false positives cannot be corrected later.

To tackle the above challenges, we further propose an online algorithm based on evolving queue of risky posts to adapt the above methods (\S \ref{sec:screening}, \S \ref{sec:HAN}) for effective and efficient early detection. First, the Risky Post Screening has already provided an efficient basis for the \textit{feature updates} step\footnote{We may view the selected posts as ``features'' in our model, since they are the actual inputs for the prediction model.}, as discussed above. However, the \textit{model inferences} remain computationally demanding. We observe that we don't have to make a prediction for each post, as some posts are not very helpful or even misleading for the detection of depression, and these posts are exactly what we would filter out with our screening approach as low risk posts. 
Therefore, we can make prediction updates only if the new post is considered 
risky enough, so that the number of model inferences can be substantially reduced. The computational costs can be further contained by limiting the number of posts used in inference, so that only the most risky and recent posts will be included.

We implement the above intuition with an evolving queue updated according to 
post risk. The whole procedure is illustrated in Figure~\ref{fig:evolving}. 
We set the capacity of the queue to $K$ to control the computational costs 
as well as the number of posts used in training. 
For each incoming post $p$, the queue is updated according to the 
following rules: 
\begin{enumerate}
    \item If the queue is not full, we will add $p$ into the queue no matter how risky it is.
    \item If the queue has already been full, we will compare the risk of $p$ with the minimum risk of all posts in the queue. If $p$ is less risky, it will not be included in the queue. Otherwise, the least risky post in the queue will be removed for the insertion of $p$. The posts in queue will then be sorted in chronological order to align with the model's positional encoding.
\end{enumerate}

The HAN model will make inference only if the queue updates to avoid unnecessary computations. If the model's predicted probability exceeds a predefined threshold, it will report a early alert of depression and stop further calculations.

%% file: experiment.tex
\section{Experiments}

In this section, we will first introduce the dataset and the compared methods. We will then further provide our models' results in the conventional depression detection setting and their efficacy and efficiency in ERD settings. We finally illustrate the explainability of the proposed method with examples.

\subsection{Dataset}

We mainly use the \textbf{eRisk2017} dataset \cite{losada2016test} in our experiment, which is adopted as the benchmark in the ERD task of CLEF 2017 \cite{losada2017erisk}. It consists of 137 depressed users and 755 control users and is divided into training/test set with 486/406 users each. The depressed users are identified with patterns like ``I was diagnosed with depression'', while the control users are those active on depression subreddit but had no depression. The posting year spans from 2007 to 2015. The anchor post for identification is filtered from the dataset. This filtering strategy can prevent the direct information leakage from the self-report, which may prevent the model from learning other indirect depression signals. We also conducted experiments and validated the generalizability of the proposed method on other two datasets with in-domain and cross-domain experiments (see Appendix). 

\subsection{Competing Methods}
We compare our method with several competitive baselines. For traditional machine learning models. \textbf{LR} uses TF-IDF features and a logistic regression classifier. \textbf{Feature-Rich} utilizes some additional user-based features, including LDA topic distribution \cite{blei2003latent}, LIWC features \cite{pennebaker2001linguistic} and emoticon counts. This is a competitive baseline that has been widely accepted in depression detection works on Facebook, Twitter and Reddit datasets \cite{eichstaedt2018facebook,trotzek2018utilizing,harrigian2020models}.

For neural baselines, we consider both models with large pretrained LM and relatively small models. For small models, we choose the representative \textbf{HAN-GRU} model \cite{zogan2021explainable}, which adopts a similar HAN structure with GRU as both the user encoder and post encoder. To have a fair competition with large models, it uses the last 1000 posts for classification, which is already a major portion of or the full posting list for many users. For large models, due to their computational cost and length limit, post selection is necessary. Therefore, each method is denoted as a pair of model and selection strategy. For backbone model, we consider the strong pre-trained model \textbf{BERT} and the proposed \textbf{HAN-BERT}. For post selection strategy, \textbf{Heuristic} chooses last posts in user history. \textbf{Clus} and \textbf{Clus+Abs} are inspired by \cite{zogan2021depressionnet}. We use sentence-bert \cite{reimers-2019-sentence-bert} to get post embeddings and run K-means clustering to get the $K$ posts nearest to the cluster center as representative posts (Clus). These posts are further passed to a BART model \cite{lewis2020bart} pretrained on CNN/DM summarization dataset to get an abstractive summary (Clus+Abs). Finally, the proposed screening strategy is denoted as \textbf{Psych}. 

The basis of BERT and HAN-BERT models are \texttt{bert-base-uncased}. The sentence-bert model is \texttt{paraphrase-MiniLM-L6-v2}. The number of selected posts is $K=16$. We train with a batch size of 4, and learning rate of 2e-5. We concatenate the selected posts as input into the BERT baselines. For HAN-BERT models, the user encoder is a 4-layer 8-head transformer encoder. To avoid the influence of randomness, we run each method with 3 different seeds and report the best performance.

\subsection{Conventional Setting Results}
\label{sec:conventional}

We first conduct experiments in conventional depression detection setting (Table \ref{table:erisk2017}). We can see that BERT (Clus+Abs) performs worse than BERT (Clus), indicating that the abstractive summarization strategy does not necessarily work possibly due to the gap between its pretrained domain (News) and Reddit. HAN-BERT (Clus) outperforms BERT (Clus), showing the effectiveness of the proposed HAN structure. The poor performance of HAN-BERT (Heuristic) highlights the importance of post selection, and none of the traditional post selection methods can outperform the competitive Feature-Rich model with access to all posts. However, with our proposed screening strategy, the HAN-BERT (Psych) model significantly outperforms baselines. HAN-GRU performs worse than HAN-BERT (Psych), suggesting the importance of a strong backbone model.

\begin{table}[t]
    \centering
    \small
    \begin{tabular}{l|c}
        \hline
        {} & F1 \\
        \hline
        LR & 60.2 \\
        Feature-Rich & 63 \\
        \hline
        HAN-GRU & 61.7 \\
        BERT (Clus) & 59.6 \\
        BERT (Clus+Abs) & 52.3 \\
        HAN-BERT (Heuristic) & 43.2 \\
        HAN-BERT (Clus) & 62.5 \\
        \hline
        HAN-BERT (Psych) & \textbf{70.3} \\
        \hline
    \end{tabular}
    \caption{\label{table:erisk2017} Results on eRisk2017 test set.}
\end{table}

\subsection{Early Detection}

We then test model performance in the ERD setting, using the official metrics ERDE$_{5}$ and ERDE$_{50}$ \cite{losada2017erisk}. We also report F1 calculated using the early predictions (report positive if predicted probability over the predefined threshold 0.5, with which all these models are trained). We exclude baselines with unsatisfying performance and preserves LR, Feature-Rich and HAN-GRU. To tackle the item-by-item updates, the baseline models have to recalculate the features and run inference for each new post, while our HAN-BERT with risky post screening can deal this efficiently with the proposed evolving queue algorithm (\S \ref{sec:evolving}). When counting the running time, we accumulate the time costs for all posts no matter if the model makes early predictions to rule out the influence of early false positives. LR and Feature-Rich run on a Linux machine with CPU: E5-2678 (48 cores), while HAN-GRU and HAN-BERT run with a NVIDIA 2080 Ti GPU. Although the comparison is not totally fair, we think it is still a practical setting for real world applications.

\begin{table}[h]
    \centering
	\small
    \begin{tabular}{l|cccccc}
        \hline
        Model & ERDE$_5$ & ERDE$_{50}$ & F1 & Time(s) \\
        \hline
        LR & 13.70 & 8.49 & 40.5 & 4710.7 \\
        Feature-Rich & 12.98 & 8.39 & 35.8 & 7558.4 \\
        HAN-GRU & 13.30 & 8.58 & 40.7 & 19671.4 \\
        HAN-BERT (Psych) & \textbf{10.72} & \textbf{8.12} & \textbf{60.3} & \textbf{1330.3} \\
        \hline
    \end{tabular}
    \caption{\label{table:early} Test results on eRisk2017 in early detection setting. The lower ERDE$_5$ and ERDE$_{50}$, the better model performs early detection.}
\end{table}

The results are shown in Table \ref{table:early}. It can be seen that HAN-BERT (Psych) significantly outperforms baselines in ERDE and F1, while also being faster. The reason for the superiority on effectiveness is because baselines produce much more false positives than HAN-BERT due to their sensitivity to single posts. The advantage of efficiency can be mainly attributed to the evolving queue algorithm, which greatly reduced the number of model inference to only 10.41\% of all posts. The efficient feature update also helps. Although the sentence encoding must be conducted for all posts, it only costs a small fraction of total time (114.7s out of 1330.3s). The extremely long running time of HAN-GRU further highlights the importance of the proposed algorithm, as we can expect an unaffordable time cost for the even larger BERT-based models without it. 

\begin{figure}[h]
    \centering
    \includegraphics[width=0.88\columnwidth]{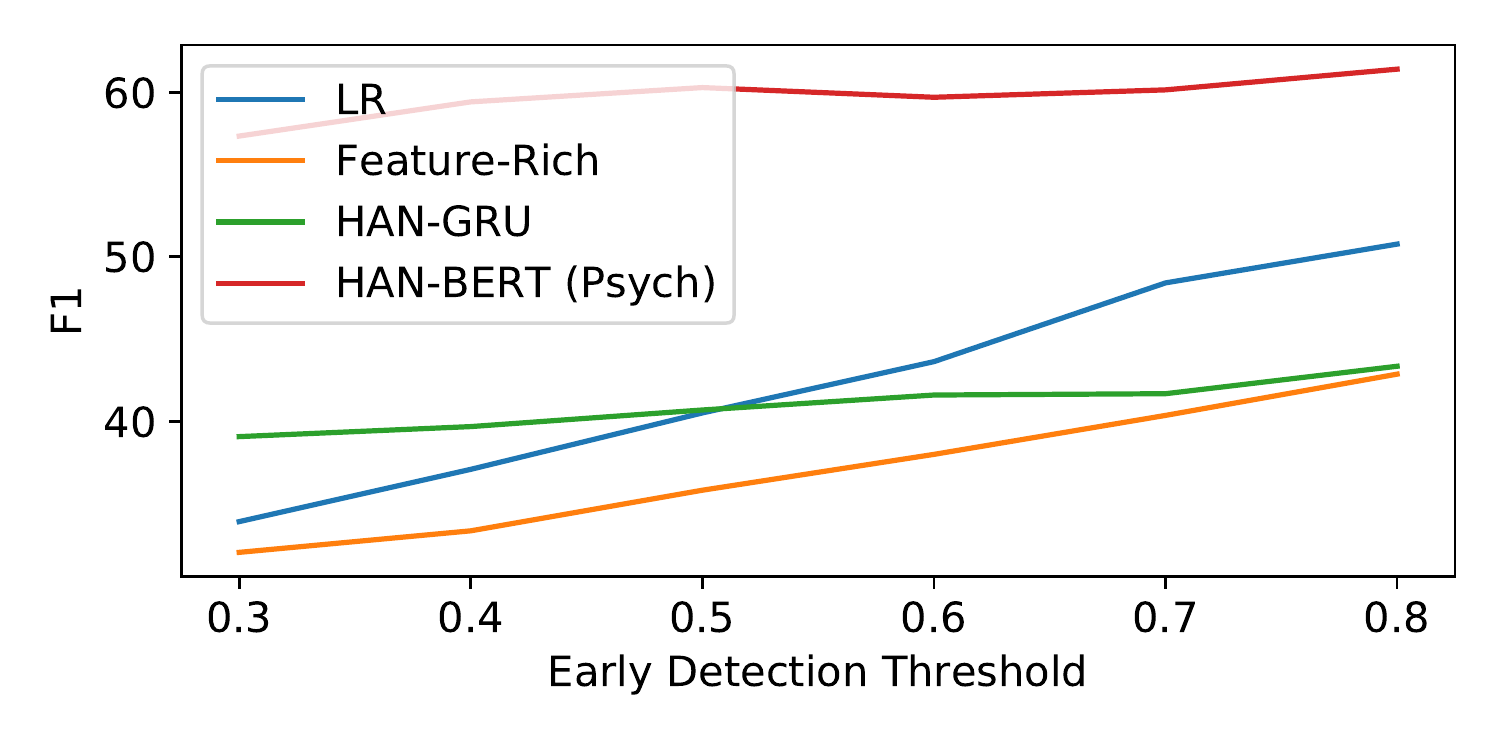}
    \caption{Effect of threshold on early detection F1.}
    \label{fig:thr2early_f1}
\end{figure}

\begin{table*}[t]
    \centering
	\small
    \begin{tabular}{ccc}
        \hline
        Attention & Post & Diagnostic Basis \\
        \hline
        0.202 & \makecell[l]{It sucks that the Citalopram didn't work for you but glad to hear your other \\ meds are helping. It's my first time on antidepressants so I didn't know \\ what are their side effects.} & \makecell[c]{\textbf{Treatment} \\ Diagnosis \\ Changes in Appetite }\\
        \hline
        0.118 & \makecell[l]{Thanks! :) Sometimes it's really good to actually get the words out of  \\ me rather than internalising my feelings.} & \makecell[c]{Concentration Difficulty \\ Loss of Pleasure \\ \textbf{Self-Dislike}}\\
        \hline
        0.048 & \makecell[l]{Glad to know :) just glad I'm not working for the next couple of weeks. \\ Feel like I'm on a different planet haha.} & \makecell[c]{\textbf{Tiredness} \\ Stay still \\ Concentrate} \\
        \hline
        0.021 & \makecell[l]{Some films or TV shows. I remember watching ... The worst part was I'd  \\ already been laughed at by my mum for crying at the end of Breakfast at \\ Tiffanys (who leaves a cat out in the rain like that?). } & \makecell[c]{Sadness \\ \textbf{Crying} \\ Depressed Mood} \\
        \hline
        \end{tabular}
        \caption{\label{table:example} Example posting list (4 selected out of all 16 posts) of a user with depression with their attention weight in HAN and diagnostic basis according to top 3 cosine similarity (reasonable ones highlighted in bold).}
\end{table*}

We can adjust the detection latency $t$ by tuning the threshold and balance the tradeoff between precision and recall. Therefore, we hypothesize that model performance can be improved with varied threshold. We tune the threshold from 0.3 to 0.8, and check the changes in their early detection F1. This will run the risk of overfitting on the test set, but allow us to explore the best possible performance. As is shown in Figure \ref{fig:thr2early_f1}, the performance of baseline systems can improve by changing the threshold, but still fall behind HAN-BERT (Psych). Moreover, the performance of HAN-BERT (Psych) is not sensitive to threshold, so we may deploy it more comfortably without concerns on threshold tuning.

\subsection{Qualitative Example}

We provide a concrete example in Table \ref{table:example} to analyze the behaviors of HAN-BERT (Psych) in detail. From the column of attention weight, we can see that posts with strong depression indicators (e.g. antidepressants, internalizing feelings, see Row 1, 2) received much higher attention than a uniform baseline (1/16 = 0.0625), while posts with no evident signals of depression or even with a positive emotion (Row 3, 4) received low attention. This suggests the usefulness of the attention weight as an explanation for model prediction. The diagnostic bases decided by the cosine similarity between post and depression templates constitute another type of intuitive explanation. The top 3 diagnostic bases can usually capture the conventional depression behaviors that the post may indicate, which may act as a convincing interpretation in its clinical applications. However, we noticed that sometimes the similarity model may rank an unreasonable aspect high in the list of bases, such as the ``Sadness'' for the last positive post. We owe such mistakes to the limitation of sentence representation models, such as not sensitive to negation \cite{ribeiro2020beyond}. We expect stronger sentence representation models to alleviate the problem.

%% file: related.tex
\section{Related Work}
Recently, depression detection has received much attention. Studies include predicting depression diagnosis from clinical interviews~\cite{gratch2014distress}, medical records~\cite{eichstaedt2018facebook} and self-reported surveys~\cite{guntuku2019twitter}. Depression detection on social media is especially promising, as proxy diagnostic signals can be relatively easy to get from self-reports or activities in depression communities~\cite{ernala2019methodological}. Early attempts by \citeauthor{losada2016test}\shortcite{losada2016test} used TF-IDF and Logistic Regression on all user posts for depression detection. Later researchers further incorporate new features like LDA, LIWC dictionary and posting patterns \cite{trotzek2018utilizing}. For deep learning methods, \citeauthor{yates2017depression}\shortcite{yates2017depression} 
uses hierarchical CNN to process all the posts of a user at the first level and merge the output at the second level for user-level classification. However, most of them directly use all the user's posts without screening out salient posts, which may negatively affect their accuracy and efficiency.

In terms of model interpretability, traditional feature-based methods are partially explainable on the level of global features. For example, \citeauthor{shen2017depression}\shortcite{shen2017depression} found different behaviors for depressed users in posting time, emotion catharsis, self-awareness and life sharing. However, these methods cannot make user-level explanations as personalized diagnostic basis. Detecting depression from its corresponding symptoms can be a promising approach to improve explainability. The pioneering work of \citeauthor{mowery2017understanding}\shortcite{mowery2017understanding} established an annotation scheme for depressive symptoms and an annotated corpus. However, the annotations are difficult so that the amount of data is not sufficient to train a reliable symptom classifier. Our approach also adopts the idea of explaining depression detection from symptoms. But it identifies symptoms implicitly with similarity matching, and thus can alleviate the requirement for large annotated corpus.

In practice, we also want to identify depression risk as early as possible, as is exemplified by the eRisk competitions~\cite{losada2019overview}. The majority of proposed methods can only achieve satisfying performance given almost the whole dataset, and few of them are able to make immediate response to each item update. To reduced the number of required posts, \citeauthor{zogan2021depressionnet}\shortcite{zogan2021depressionnet} uses extractive summarization to extract key posts of a user. However, it relies on K-means clustering to get the summaries, so the model cannot run online as well.

%% file: conclusion.tex
\section{Conclusions}

In this work, we tackle the problem of ERD of depression detection with a novel, psychiatry-guided method of risky post screening and hierarchical attentional network. The accurate selection of risky posts out of the long user history constitutes a solid foundation for prediction as well as enables the usage of large pretrained language model. Furthermore, our framework can work on ERD scenarios with high efficiency, supported by the proposed evolving queue algorithm, which can greatly reduce the required number of model inferences. Utilizing attention mechanism and depression scales provides our method with strong interpretability in the form of attention weights and diagnostic basis, which we hope can facilitate its further application in online detection as a reliable assistant.

%% file: ethical.tex
\section{Ethical Statement}

The datasets used in this work are either publicly available or used under their corresponding data usage agreement. All posts in examples were de-identified and paraphrased for anonymity. We provide further discussion in Appendix. 